\documentclass[11pt,preprint]{declare}

\usepackage[absolute,overlay]{textpos}
\usepackage{tikz}
\usepackage{hyperref}
\usepackage{url}
\usepackage{amsthm}
\usepackage[table]{xcolor}

\usepackage{hyperref}

\usepackage{fontawesome5}
\usepackage[smartEllipses]{markdown}
\usepackage{pifont}
\usepackage{xcolor}
\usepackage{bbm}
\usepackage[table]{xcolor}
\usepackage{multirow}
\usepackage{subcaption}
\usepackage{wrapfig}
\usepackage{xspace}

\usepackage{wrapfig}
\usepackage{multirow}

\usepackage{graphicx,xcolor,float}

\usepackage{threeparttable}

\usepackage[ruled,vlined]{algorithm2e}
\usepackage{colortbl}
\usepackage{color}
\usepackage{multirow}
\usepackage{tabularx}
\usepackage{float}
\usepackage{amsmath, amssymb}
\usepackage{tikz}
\usetikzlibrary{shapes.geometric, arrows.meta, positioning, calc, fit, backgrounds}
\usepackage{graphicx}
\usepackage{booktabs}
\usepackage{arydshln}
\usepackage{enumitem}
\usepackage{wrapfig}
\usepackage{caption}
\usepackage{graphicx}
\usepackage{makecell}
\usepackage{float} 
\usepackage{soul}
\usepackage{mathtools}

\usepackage{bbding}

\usepackage{makecell}
\usepackage{tabularx}
\usepackage{amssymb,mathrsfs,amsmath}

\usepackage{pifont}
\usepackage{xcolor}
\usepackage{bbm}
\usepackage[table]{xcolor}
\usepackage{multirow}

\usepackage{xspace}

\usepackage{fontawesome5}

\usepackage{listings}

\runningtitle{Pala et al. (2026). GRAIL: Gradient-Reweighted Advantages for Reinforcement Learning with Verifiable Rewards.}

\runningstatus{Under review}

\usepackage[table,xcdraw,usenames,dvipsnames]{xcolor}

\usepackage{cleveref}
\SetKwInOut{Input}{Input}\SetKwInOut{Output}{Output}
\usepackage{twemojis}

\hypersetup{
    colorlinks=true,
    linkcolor=red,
    citecolor=cyan,
    filecolor=magenta,      
    urlcolor=magenta,
    }

\declarehomelogos[10.64mm]{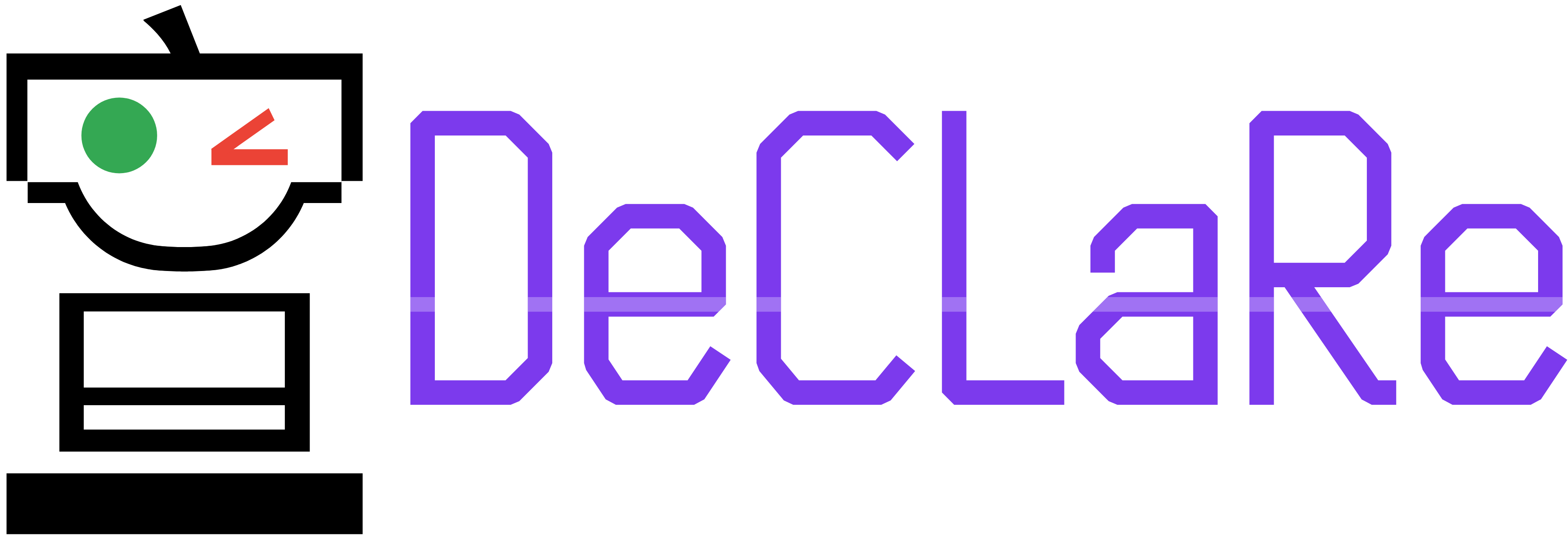, 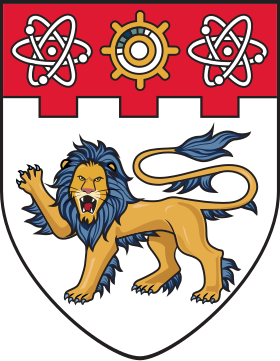}
\title{\method{}: Gradient-Reweighted Advantages for Reinforcement Learning with Verifiable Rewards}

\author[1]{Tej Deep Pala}
\author[1]{Vernon Toh}
\author[1]{Soujanya Poria}
\affiliation[1]{\small{DeCLaRe Lab, Nanyang Technological University}}

\github{grail}

\correspondence{Soujanya Poria (\email{soujanya.poria@ntu.edu.sg})}
\definecolor{forestgreen}{RGB}{34,139,34}

\newcommand{\method}{\texttt{GRAIL}\xspace}
\newcommand{\grpo}[1]{\texttt{GRPO}}
\newcommand{\oar}[1]{\texttt{OAR-G}}

\abstract{
Reinforcement learning with verifiable rewards (e.g., GRPO) is now a common way to improve mathematical reasoning in large language models (LLMs). However, current methods usually broadcast one sequence-level advantage to all tokens, or use costly process reward models (PRMs) for step-level supervision. Uniform advantage distribution assumes that all tokens contribute equally to the final reward. This dilutes the gradient signal, since flawed reasoning steps and filler words are updated as strongly as valid logical inferences. To address this, we introduce Gradient-Reweighted Advantage (\method{}), an intrinsic token-wise advantage reweighting method. \method{} uses gradient-activation saliency to place more weight on tokens that are more locally sensitive to the final answer. Evaluations across five models from the Qwen3, R1-distilled and OctoThinker families show that \method{} consistently outperforms \grpo{}. \method{} achieved an average improvement of 3.60\% in accuracy and 3.05\% in Pass@3, demonstrating that fine-grained reasoning alignment can be achieved without process-level supervision.
}

\date{\today}

\begin{document}

\maketitle

\section{Introduction}

Enhancing mathematical reasoning capabilities in large language models (LLMs) has become a critical research frontier. For domains where correctness can be programmatically verified, reinforcement learning with verifiable rewards has emerged as a primary paradigm. Traditionally, alignment has relied on algorithms like PPO~\citep{schulman2017proximal} to optimise reward models based on human feedback; however, the focus has recently shifted toward more memory-efficient alternatives that utilise outcome-based reward signals to cultivate complex reasoning behaviours without necessitating a separate critic model.

Recent literature highlights the efficacy of this outcome-driven approach. Group Relative Policy Optimization (\grpo{}) introduced a framework by normalising rewards within a group of sampled completions~\citep{shao2024deepseekmath}. At scale, GRPO-style training has produced strong chain-of-thought reasoning models, such as DeepSeek-R1~\citep{guo2025deepseek} and Qwen-Math~\citep{yang2024qwen2}. Subsequent studies have sought to refine this approach; for instance, DAPO addresses instabilities from entropy collapse~\citep{yu2026dapo}, while BNPO~\citep{xiao2025bnpo} and Dr.\ GRPO~\citep{liu2025understanding} propose alternative normalisation strategies to mitigate biases from variable completion lengths.

Despite these advances, credit assignment across the reasoning trace remains coarse. Current GRPO-style methods usually broadcast the same sequence-level advantage to all tokens in a completion. This assumes that all tokens contribute equally to the final outcome. As a result, the gradient signal is diluted. Flawed reasoning steps and filler words can receive updates as strong as critical logical steps. We hypothesise that reweighting the token-level advantage with an intrinsic saliency signal can focus the update and improve reasoning alignment.
\begin{figure*}[t]
    \centering
    \includegraphics[width=0.9\textwidth]{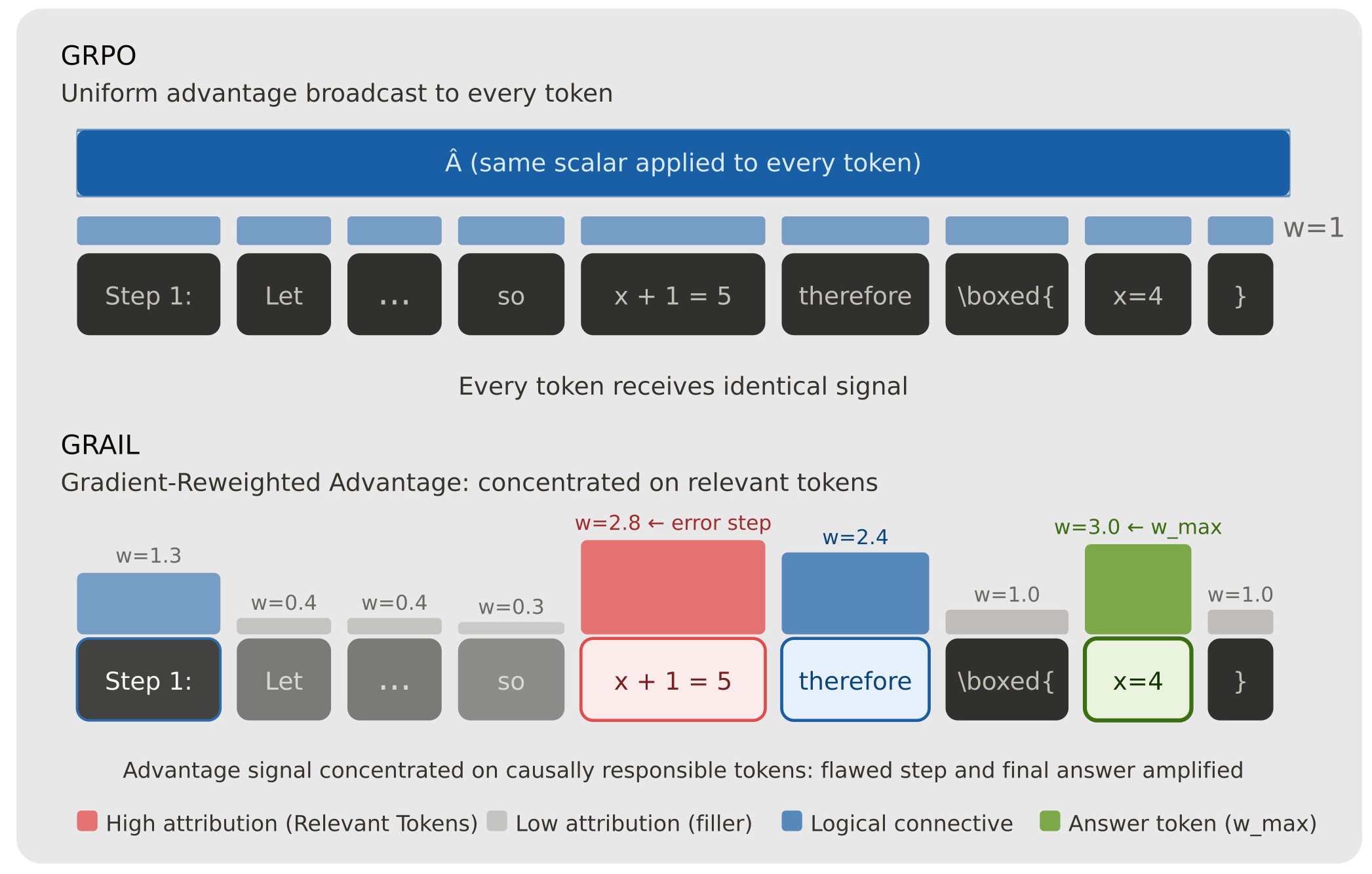}
    \caption{Comparison of credit assignment in \grpo{} versus \method{}. \textbf{Top:} Standard \grpo{} broadcasts a uniform scalar advantage across all tokens in a rollout. \textbf{Bottom:} \method{} uses gradient-activation saliency to reweight token-level advantages. High-saliency tokens receive stronger updates, while incidental filler words receive weaker updates.}
    \label{fig:intuition}
    \vspace{-10pt}
\end{figure*}

To address this limitation, this paper introduces Gradient-Reweighted Advantage (\method{}), an intrinsic token-wise advantage reweighting method. \method{} computes a gradient-activation saliency score for each token. It then uses this score to reweight the sequence-level advantage in the policy-gradient loss. Extensive evaluations across five models from the Qwen3, R1-distilled and OctoThinker families show that \method{} consistently outperforms the \grpo{} baseline, with an improvement of 3.60\% in average accuracy and 3.05\% in Pass@3.

\section{Related Work}
\label{sec:related}

\subsection{Process Supervision and Reward Models}
Outcome reward models (ORMs) provide a single binary signal at the end of a completion, offering no guidance on which reasoning steps were correct or incorrect. Process reward models (PRMs) address this by assigning a reward to each intermediate reasoning step, providing a denser supervision signal that can identify where in a chain-of-thought an error was introduced. While PRMs trained on human labels \citep{lightman2024let} or via automated verification like Math-Shepherd \citep{wang-etal-2024-math} outperform ORMs on complex benchmarks, they remain difficult to scale. A key limitation of PRM-based approaches is that they require extensive human annotation or rely on trained verifiers, and automatically synthesised data often yields inferior performance~\citep{zhang2025lessons}. In contrast, \method{} derives token-level weights directly from the model's intrinsic gradients during training, providing fine-grained credit assignment without the overhead of external supervision.

\subsection{Gradient-Based Saliency}

Saliency methods estimate which input features are most important for a model output. Gradient-based methods use the gradient of the output with respect to the input as a local sensitivity signal \citep{baehrens2010explain, simonyan2013deep, selvaraju2020grad}. The element-wise gradient $\times$ activation estimator used by \method{} is a first-order Taylor-style approximation \citep{baehrens2010explain}. It is cheaper than path-integral methods such as Integrated Gradients \citep{sundararajan2017axiomatic}.

In language models, these methods are often used after training to identify influential tokens~\citep{bastings2020elephant, ferrando-etal-2022-measuring} or to trace factual knowledge to model components~\citep{meng2022locating}. Attention weights are also used as an importance proxy, but they are not a reliable indicator of causal effect~\citep{clark-etal-2019-bert, jain2019attention, wiegreffe2019attention}. Using saliency during training is less explored. Some token-MDP methods assign per-step rewards by marginalising over future trajectories \citep{arjona2019rudder, pignatelli2023survey, chan2024dense}. \method{} is different. It uses gradient-based saliency to reweight the existing \grpo{} objective, without changing the reward function or rollout procedure.

\subsection{Fine-Grained Credit Assignment and Advantage Reweighting}

To address the coarse nature of sequence-level advantage in algorithms like GRPO, recent works have explored mechanisms to redistribute credit at the token level. Several approaches use intrinsic uncertainty metrics, such as token entropy, to scale advantages. For instance, \citet{cheng2026reasoning} amplify advantages on high-entropy tokens to encourage exploration, while \citet{chen2025beyond} segment generations by entropy and reweight advantages based on the overlap of low-entropy segments between correct and incorrect rollouts. Other heuristics include upweighting critical tokens based on attention dynamics \citep{li2025attention} or reshaping advantages using a combination of response-level confidence and token-level logit certainty \citep{xie2025unlocking}. 

Moving beyond heuristic proxies, PRIME \citep{cui2025process} attempts to tackle the credit assignment problem by training an implicit PRM online using only outcome labels, generating dense token-level rewards that are aggregated into the Monte Carlo advantage estimate. Closely related to our work is Outcome-grounded Advantage Reshaping (OAR) \citep{li2026outcome}, which redistributes advantages using a bi-level gating mechanism based on token influence. To approximate this influence, OAR's gradient-based variant (OAR-G) calculates the KL divergence between a clean outcome distribution and a noisy outcome distribution derived by injecting Gaussian noise into the embeddings. Consequently, OAR requires two forward passes (one clean, one noisy) and one backward pass per sequence to compute the self-distillation gradient. In contrast, \method{} directly leverages the analytic gradient of the final answer loss ($\mathcal{L}_{\text{ans}}$) with respect to the input embeddings. This fundamental formulation difference allows \method{} to extract a precise attribution signal using only a single forward and a single backward pass, eliminating the computational overhead of noise injection and auxiliary self-distillation objectives. Furthermore, \method{} introduces targeted structural interventions to ensure valid weighting at delimiter and EOS tokens, ensuring a stable and mathematically direct attribution link between intermediate reasoning and the final outcome.
 
\begin{figure*}[t]
    \centering
    \fbox{ \includegraphics[width=0.8\textwidth]{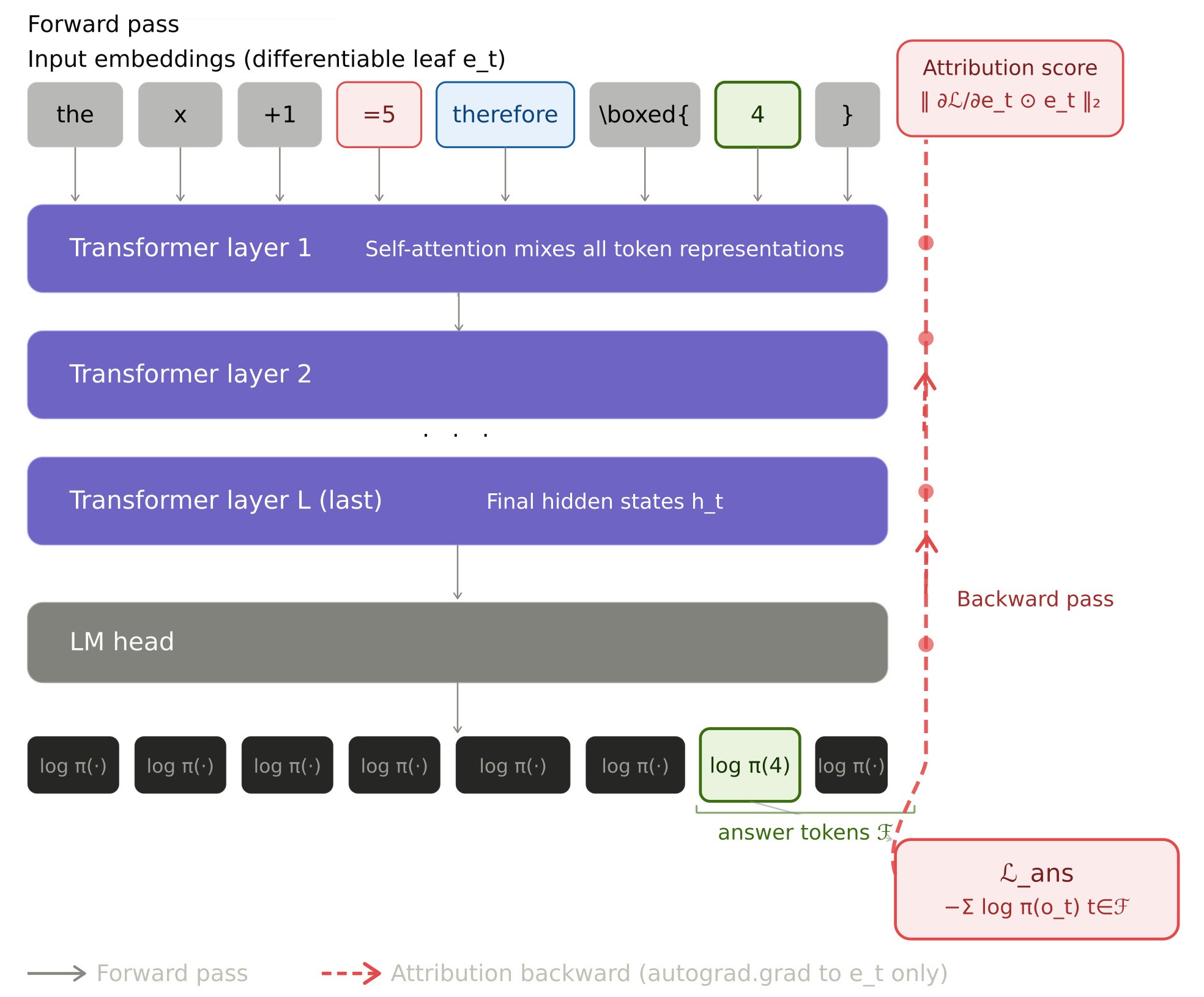}}
    \caption{The gradient-activation saliency mechanism in \method{}. To compute the token-specific saliency score, the differentiable leaf is placed at the input embeddings ($\mathbf{e}_t$). The gradient of the answer loss ($\mathcal{L}_{\text{ans}}$) flows back through the sequence-wide self-attention layers. This gives a local sensitivity signal for earlier reasoning tokens with respect to the final answer tokens.}
    \label{fig:architecture}
    \vspace{-10pt}
\end{figure*}

\section{Methodology}

Reinforcement learning from verifiable rewards usually spreads one advantage over the whole generated sequence. This ignores that some reasoning tokens matter more than others. Our goal is to make this credit assignment less noisy. In this section, we introduce Gradient-Reweighted Advantage (\method{}), an intrinsic token-wise advantage reweighting method. \method{} uses gradient-activation saliency to reweight the sequence-level advantage before applying the policy-gradient update.

\subsection{Token-Wise Advantage Reweighting}

Consider a language model policy parameterised by $\theta$, denoted as $\pi_\theta$. Given a query $q$, the model generates an output consisting of a sequence of tokens $o = \{o_1, o_2, \dots, o_T\}$. In the context of Group Relative Policy Optimization (\grpo{}), a group of rollouts $G$ is generated for the same query, and an outcome-based reward $r_i$ is computed to derive a relative advantage estimate $\hat{A}_i$ for each rollout $o_i$. In our experiments, we use the Dr.\ GRPO style of advantage computation:

$$A_i = r_i - \frac{1}{|G|}\sum_{i=1}^{|G|}{r_i}$$

To simplify notation, we define the importance sampling ratio of the active policy to the reference policy at timestep $t$ as $\rho_{i,t}(\theta)$:

\begin{equation}
\rho_{i,t}(\theta) = \frac{\pi_\theta(o_{i,t} \mid q, o_{i,<t})}{\pi_{\theta_{old}}(o_{i,t} \mid q, o_{i,<t})}
\end{equation}

The standard policy gradient objective for \grpo{} (excluding the KL divergence penalty) is formulated as:

\begin{align*}
\mathcal{L}_{\text{GRPO}} =&  -\frac{1}{|G|} \sum_{i=1}^{|G|} \frac{1}{|o_i|} \sum_{t=1}^{|o_i|} \min \Big( \rho_{i,t}(\theta) \hat{A}_i, \text{clip}\big(\rho_{i,t}(\theta), 1-\epsilon, 1+\epsilon\big) \hat{A}_i \Big)
\end{align*}

A limitation of this formulation is the identical use of the sequence-level advantage $\hat{A}_i$ across all tokens at timestep $t$. Under this paradigm, a filler word or a flawed intermediary derivation that coincidentally precedes a correct answer receives the same gradient update as the correct logical inferences that actively drove the positive outcome. The model is effectively unable to distinguish which specific parts of its reasoning chain to correct or reinforce.

To resolve this, we introduce a token-specific saliency weight, $w_{i,t}$. This weight is a local proxy for how sensitive the final answer is to token $o_{i,t}$. We integrate it directly into the policy-gradient objective:

\begin{align*}
\mathcal{L}_{\text{GRAIL}} =&  -\frac{1}{|G|} \sum_{i=1}^{|G|} \frac{1}{|o_i|} \sum_{t=1}^{|o_i|} \min \Big( \rho_{i,t}(\theta) \text{sg}(w_{i,t}) \hat{A}_i,  \text{clip}\big(\rho_{i,t}(\theta), 1-\epsilon, 1+\epsilon\big) \text{sg}(w_{i,t})\hat{A}_i \Big)
\end{align*}

\subsection{Gradient-Activation Saliency}

We need a token-specific saliency score to compute $w_{i,t}$. We use a gradient-activation saliency method. To get a non-zero signal for intermediate reasoning tokens, we compute the gradient with respect to a representation before the transformer body. This lets the backward pass propagate through all causal attention layers. We therefore use the input embeddings, $\mathbf{e}_t$, as the differentiable leaf.

Let $\mathcal{F}$ denote the subset of positional indices corresponding to the final answer tokens (e.g., the tokens constrained within a \texttt{\textbackslash boxed{\dots}} final answer marker). The outcome-based loss, $\mathcal{L}_{\text{ans}}$, is defined as the negative log-likelihood of these answer tokens:

$$\mathcal{L}_{\text{ans}} = - \sum_{t \in \mathcal{F}} \log \pi_\theta(o_t \mid o_{<t})$$

The gradient flow from the answer loss backward to the input embedding of an arbitrary token $t$ is continuously mediated by the self-attention mechanism across all $L$ layers of the network. For any preceding reasoning token $r \notin \mathcal{F}$, the gradient $\frac{\partial \mathcal{L}_{\text{ans}}}{\partial \mathbf{e}_r}$ evaluates to a non-zero vector exclusively when the attention mechanism has routed information from position $r$ forward into the final hidden states of the answer tokens, $\mathbf{h}_{\mathcal{F}}^{(L)}$.

We define the raw saliency score, $s_t$, for each token as the $L_2$-norm of the element-wise product between the gradient of the answer loss with respect to its input embedding and the embedding itself:

$$s_t = \left\| \frac{\partial \mathcal{L}_{\text{ans}}}{\partial \mathbf{e}_t} \odot \mathbf{e}_t \right\|_2$$

Empirically, the raw saliency scores $s_t$ are non-negative and exhibit a heavy-tailed distribution. Consequently, a small subset of tokens yields scores several orders of magnitude larger than the bulk of the sequence. Applying standard z-score normalisation directly to these raw scores is suboptimal, as the resulting weights would be disproportionately dominated by extreme outliers, leading to erratic gradient updates.

To mitigate this and ensure a stable learning signal, we project the saliency scores into log-space prior to standardisation. We define the log-score $\tilde{s}_t$ as:

$$\tilde{s}_t = \log(s_t + \epsilon)$$

where $\epsilon$ is a small constant added to ensure numerical stability. The final saliency weight $w_{t}$ is then computed by standardising these log-scores and scaling them to a controlled distribution:

$$w_t = w_{\text{mean}} + \frac{\tilde{s}_t - \mu_{\log}}{\sigma_{\log}} \cdot \sigma_w$$

$\mu_{\log}$ and $\sigma_{\log}$ denote the mean and standard deviation of the log-scores across all completion tokens in the rollout. The hyperparameter $w_{\text{mean}}$ establishes the baseline weight for a neutral token, while $\sigma_w$ controls the spread and intensity of the saliency reweighting.

Finally, to guarantee training stability and prevent any single high-saliency token from overwhelmingly dominating the policy gradient, the derived weights are strictly clipped to a bounded range $[w_{\min}, w_{\max}]$.

This gives a first-order local sensitivity signal for each token with respect to the final answer. The score is computed from the current model during training. It does not require external heuristics or an independently trained process reward model.

Furthermore, the saliency weights are strictly wrapped in a stop-gradient operator, $\text{sg}(\cdot)$. This is essential because the weights themselves depend on $\theta$ through the saliency computation. Without the stop-gradient, the optimiser would encounter an unaccounted second-order term, $\frac{\partial w_t}{\partial \theta}$, rendering the objective ill-defined as a standard policy gradient estimator. By treating the weights as fixed scalars from the optimiser's perspective, $\mathcal{L}_{\text{GRAIL}}$ remains a valid, importance-weighted policy gradient.

\subsection{Simple Weight Corrections}

A few token types need simple corrections. First, the final answer tokens often receive low saliency scores, because their gradient signal is mostly self-referential. To keep final correctness important, we set the weights at these inner answer positions to the maximum bound, $w_{\max}$. In contrast, final answer delimiters, such as the opening \texttt{\textbackslash boxed\{}, can receive high scores because they are close to the answer. We do not want to reinforce formatting over reasoning, so we assign delimiter positions a neutral weight of $w_{\text{mean}}$. 

Finally, tokens generated after the final answer, including the EOS token, have near-zero saliency. They do not have a differentiable path to the answer loss under the causal attention mask. If we suppress these updates, the model may become less certain about when to stop. We therefore reset all post-answer positions to $w_{\text{mean}}$.

\section{Experimental Setup}

To rigorously evaluate the efficacy of Gradient-Reweighted Advantage (\method{}) against Group Relative Policy Optimization (\grpo{}) baseline, we design a comprehensive experimental framework targeting mathematical reasoning capabilities.

\subsection{Training}

Our experiments are conducted across multiple architectures to ensure the generalisability of the proposed saliency mechanism. Specifically, we utilise five models drawn from three distinct families: Qwen3 (4B and 8B) \citep{qwen3technicalreport}, Deepseek-R1-Distill-Llama-8B \citep{guo2025deepseek} and OctoThinker (OctoThinker-3B-Short and OctoThinker-8B-Short) \citep{wang2025octothinker}. All models are then trained on the DeepMath-103K dataset \citep{he2025deepmath}, which provides a robust and diverse corpus of verifiable mathematical problems suitable for outcome-based reinforcement learning.

To test the impact of our loss reweighting formulation, the base models, datasets, and core training infrastructure remain identical across both the \method{} and \grpo{} runs. A comprehensive breakdown of all training hyperparameters, including learning rates, global batch sizes, and \method{} hyperparameters, is detailed in Appendix \ref{sec:hyperparam}.

\subsection{Evaluation}

We evaluate the generalisation of mathematical problem-solving capabilities across six established benchmarks encompassing varying degrees of difficulty: Math500 \citep{lightman2024let}, AIME 2024 \citep{aime24}, AMC 2023 \citep{mathai_amc23}, MinervaMath \citep{lewkowycz2022solving}, CollegeMath \citep{tang2024mathscale}, and OlympiadBench \citep{he2024olympiadbench}.

\begin{table*}[t]
\centering
\resizebox{1.0\textwidth}{!}{%
\begin{tabular}{lccccccc}
\toprule
\textbf{Model} &
\textbf{MATH500} &
\textbf{AMC23} &
\textbf{AIME24} &
\textbf{College MATH} &
\textbf{Olympiad Bench} &
\textbf{Minerva MATH} &
\textbf{Average} \\
\midrule
& \multicolumn{7}{c}{\textit{Average Accuracy}} \\
\midrule
Qwen3-4B & & & & & & &  \\
\quad \grpo{}  & 88.07  & 77.50  & 27.78  & 52.17 & 45.83 & 39.83 & 55.20\\
\quad \method{} & 91.47 & 80.00 & 36.67  & 53.17         & 47.50         & 44.36         & \textbf{58.86} \textcolor{forestgreen}{(+3.66)}\\
\midrule
Qwen3-8B & & & & & & &  \\
\quad \grpo{} & 90.07  & 77.50  & 36.67  & 54.00         & 46.67         & 41.42         & 57.72   \\
\quad \method{} & 92.07 & 82.50 & 47.78  & 56.00         & 49.17         & 44.73         & \textbf{62.04} \textcolor{forestgreen}{(+4.32)} \\
\midrule
OctoThinker-3B & & & & & & &  \\
\quad \grpo{} & 43.87  & 12.50  & 1.11   & 28.17         & 15.00         & 14.95         & 19.27   \\
\quad \method{} & 46.40  & 26.67  & 3.33   & 28.83         & 16.00         & 16.05         & \textbf{22.88}\textcolor{forestgreen}{(+3.61)}  \\
\midrule
OctoThinker-8B & & & & & & &  \\
\quad \grpo{} & 48.53  & 22.50  & 2.22   & 29.00         & 14.67         & 21.08         & 23.00   \\
\quad \method{} & 54.53  & 23.33  & 4.44   & 33.00         & 20.50         & 25.12         & \textbf{26.82} \textcolor{forestgreen}{(+3.82)}  \\
\midrule
R1-Distill-Llama-8B  & & & & & & &  \\
\quad \grpo{} & 81.00 &	71.67 &	21.11 &	42.50 &	42.17 &	30.64 & 48.18 \\
\quad \method{} &	82.60 &	81.67 &	25.56 &	41.67 &	42.00 &	30.88 & \textbf{50.73} \textcolor{forestgreen}{(+2.55)} \\
\midrule
& \multicolumn{7}{c}{\textit{Pass@3}} \\
\midrule
Qwen3-4B & & & & & & &  \\
\quad \grpo{}  & 93.60  & 92.50  & 43.33  & 55.50         & 53.50         & 47.06         & 64.25 \\
\quad \method{} & 93.80  & 97.50  & 53.33  & 55.00         & 54.00         & 47.79         & \textbf{66.90} \textcolor{forestgreen}{(+2.65)}  \\
\midrule
Qwen3-8B & & & & & & &  \\
\quad \grpo{} & 95.00  & 90.00  & 50.00  & 57.00         & 52.50         & 45.96         & 65.08 \\
\quad \method{} & 95.40  & 87.50  & 60  & 59.00         & 53.00         & 48.89         & \textbf{67.29} \textcolor{forestgreen}{(+2.21)}   \\
\midrule
OctoThinker-3B & & & & & & &  \\
\quad \grpo{} & 59.60  & 22.50  & 3.33   & 38.50         & 26.00         & 24.63         & 29.09   \\
\quad \method{} & 61.40  & 40.00  & 6.67   & 39.00         & 27.00         & 24.63         & \textbf{33.12} \textcolor{forestgreen}{(+4.03)}   \\
\midrule
OctoThinker-8B & & & & & & &  \\
\quad \grpo{} & 65.20  & 32.50  & 3.33   & 38.50         & 25.50         & 31.99         & 32.84   \\
\quad \method{} & 68.40  & 40.00  & 10.00  & 43.50         & 29.50         & 36.40         & \textbf{37.97} \textcolor{forestgreen}{(+5.13)}  \\
\midrule
R1-Distill-Llama-8B  & & & & & & &  \\
\quad \grpo{} & 89.00 &	85.00 &	36.67 &	50.00 &	48.50 &	42.28 & 58.57 \\
\quad \method{} & 91.00 & 90.00 & 36.67 & 49.50 &	50.00 &	41.54 & \textbf{59.79} \textcolor{forestgreen}{(+1.22)} \\
\bottomrule
\end{tabular}
}
\caption{Average accuracy and Pass@3 (\%) across six mathematical reasoning benchmarks. Results are evaluated at a sampling temperature of 0.6 and averaged over three generations per problem. Across all evaluated models, \method{} consistently achieves higher overall reasoning performance than the \grpo{} baseline.}
\label{tab:main_results}
\end{table*}

During inference, we adopt a consistent sampling strategy across all evaluation datasets. For each problem, we sample three distinct generations using a sampling temperature of 0.6. A list of evaluation hyperparameters is provided in Appendix \ref{sec:hyperparam}. Final model performance is reported as the pass rate and average accuracy across the three generations. This ensures that the reported metrics reflect the model's reliable reasoning capacity.

\section{Results}

In this section, we evaluate the efficacy of Gradient-Reweighted Advantage (\method{}) across five model architectures and six mathematical reasoning benchmarks. The empirical results demonstrate that dynamically redistributing the gradient signal to influential tokens consistently elevates problem-solving capabilities, yielding an average improvement of 3.60\% in accuracy and 3.05\% in Pass@3 over the \grpo{} baseline.

\subsection{Main Results}

Table \ref{tab:main_results} details the performance of the Qwen3 and OctoThinker model families across the evaluation suite. Overall, \method{} yields an impressive average absolute improvement of approximately 3.60\% in Average Accuracy and 3.05\% in Pass@3 across all evaluated models.

Most notably, \method{} drives substantial gains on the most challenging benchmarks that require extended, complex reasoning chains. On the AIME 2024 benchmark, Qwen3-4B improves its accuracy from 27.78\% to 36.67\%, while Qwen3-8B leaps from 36.67\% to 47.78\%. Similarly, OctoThinker-8B doubles its accuracy (2.22\% to 4.44\%) and triples its Pass@3 rate (3.33\% to 10.00\%) on the same benchmark. The Qwen3-8B architecture exhibits the largest overall benefit in average accuracy, recording a 4.32\% absolute boost, while OctoThinker-8B sees the highest Pass@3 increase at 5.13\%. This suggests that the gradient-activation saliency mechanism is highly effective at unlocking reasoning bottlenecks regardless of the model's base reasoning capacity.

Compared to \grpo{}, the targeted reinforcement in \method{} successfully amplifies critical logical steps across the entirety of the evaluation suite without any task-specific degradation. Whether evaluated on complex, multi-step competition mathematics (e.g., AIME and AMC) or advanced curriculum-level benchmarks, intrinsic loss reweighting consistently elevates performance. This confirms that the method generalises well across different mathematical problem-solving distributions.

To verify the statistical significance of the observed performance gains, we conducted a Wilcoxon signed-rank test on the paired average accuracy scores between the \grpo{} baseline and \method{} across all evaluated models and benchmarks. The test shows a statistically significant improvement when using \method{} ($W = 459.5$, $p = 1.3 \times10^{-8}$). This suggests that the gains are unlikely to come from random variation alone.

These findings show that token-wise advantage reweighting can improve reasoning alignment without step-level supervision.

\subsection{Comparison with \oar{}}
\label{sec:oar_comparison}

\begin{table*}[ht]
\centering
\resizebox{1.0\textwidth}{!}{%
\begin{tabular}{lccccccc}
\toprule
\textbf{Model} &
\textbf{MATH500} &
\textbf{AMC23} &
\textbf{AIME24} &
\textbf{College MATH} &
\textbf{Olympiad Bench} &
\textbf{Minerva MATH} &
\textbf{Average} \\
\midrule
Qwen3-4B & & & & & & &  \\
\quad \oar{} & 	89.73 &	81.67 &	30.00 &	53.50 &	46.67 & 42.77 & 57.39 \\
\quad \method{} & 91.47 & 80.00 & 36.67  & 53.17         & 47.50         & 44.36         & \textbf{58.86} \textcolor{forestgreen}{(+1.47)} \\
\midrule
Qwen3-8B & & & & & & &  \\
\quad \oar{} & 	89.27 & 74.17 & 32.22 & 53.33 &	46.67 & 38.85 & 55.75 \\
\quad \method{} & 92.07 & 82.50 & 47.78  & 56.00         & 49.17         & 44.73         & \textbf{62.04} \textcolor{forestgreen}{(+6.29)} \\
\bottomrule
\end{tabular}
}
\caption{Performance comparison between \method{} and \oar{}. \method{} achieves higher average accuracy across both architectures, with a +6.29\% absolute improvement on the Qwen3-8B model. The substantial gains on long-horizon tasks like AIME24 demonstrate that \method{}'s direct analytic gradient approach isolates critical reasoning steps more effectively than OAR's noise-injected sensitivity proxy.}
\label{tab:oar_results}
\end{table*}

Table \ref{tab:oar_results} details the performance of \method{} and \oar{} when applied to the Qwen3-4B and Qwen3-8B architectures. \method{} consistently outperforms OAR across the evaluation suite. On the 4B model, \method{} yields a 1.47\% absolute improvement in average accuracy. Crucially, as the base model scales to 8B parameters, the performance delta widens dramatically. \method{} achieves a 62.04\% average accuracy on Qwen3-8B compared to OAR's 55.75\%, representing a 6.29\% absolute gain. 

The superiority of \method{} is most pronounced on highly complex, multi-step derivation tasks. On AIME 2024, \method{} increases accuracy by 6.67\% on the 4B model and 15.56\% on the 8B model relative to OAR. While OAR relies on injecting Gaussian noise into embeddings to approximate token influence via a self-distillation objective, this proxy signal can become misaligned with strict verifier acceptance on long reasoning traces. The empirical gap between the two methods suggests that deriving the attribution signal directly from the analytic gradient of the answer loss ($\mathcal{L}_{\text{ans}}$) provides a sharper and more reliable attribution between intermediate logic and final mathematical correctness.

\subsection{Ablations}
\subsubsection{Differentiable Leaf Placement}

\begin{table*}[ht]
\centering
\resizebox{1.0\textwidth}{!}{%
\begin{tabular}{lccccccc}
\toprule
\textbf{Differentiable Leaf Layer} &
\textbf{MATH500} &
\textbf{AMC23} &
\textbf{AIME24} &
\textbf{College MATH} &
\textbf{Olympiad Bench} &
\textbf{Minerva MATH} &
\textbf{Average} \\
\midrule
Input Embeddings    & 88.67  & 81.67  & 36.67  & 51.83 & 48.17 & 41.91 & 58.15 \\
First Hidden Layer & 88.47  & 79.17  & 28.89  & 51.83         & 47.17         & 40.56 & 56.01 \\
Middle Hidden Layer & 89.13  & 76.67  & 26.67  & 52.33         & 45.33         & 40.20 & 55.05 \\
Penultimate Layer & 86.47  & 72.50  & 24.44  & 51.83 & 44.50 & 40.69 & 53.41 \\
\bottomrule
\end{tabular}
}
\caption{Ablation on the differentiable leaf placement for \method{} using Qwen3-4B. Deriving saliency scores from the input embeddings yields the highest average accuracy. Placing the leaf at progressively deeper intermediate layers steadily degrades performance, confirming that gradient propagation must capture all layers to assign credit to earlier reasoning tokens accurately.}
\label{tab:leaf_ablation}
\end{table*}

To validate the architectural design of the gradient-activation saliency, we conduct an ablation on the placement of the differentiable leaf. Table \ref{tab:leaf_ablation} details the performance of Qwen3-4B when the saliency score is derived from various depths within the model: the input embeddings, and the first, middle and penultimate hidden layers. 

Performance drops as the differentiable leaf is moved deeper into the network. Deriving saliency scores at the input embeddings yields the highest average accuracy (58.15\%). In contrast, truncating the backward pass by placing the leaf at the penultimate layer results in the lowest performance (53.41\%), alongside substantial drops on complex benchmarks like AIME24 (declining from 36.67\% to 24.44\%). 

This steady degradation confirms our theoretical hypothesis regarding credit assignment. To compute useful weights for earlier reasoning tokens, the gradient must fully propagate backwards through the entire sequence-wide attention mechanism. Bypassing these layers strips the gradient of its structural context, rendering the resulting saliency weights less effective for policy optimisation.

\begin{table*}[ht]
\centering
\resizebox{1.0\textwidth}{!}{%
\begin{tabular}{lccccccc}
\toprule
\textbf{\method{} Application Strategy} &
\textbf{MATH500} &
\textbf{AMC23} &
\textbf{AIME24} &
\textbf{College MATH} &
\textbf{Olympiad Bench} &
\textbf{Minerva MATH} &
\textbf{Average} \\
\midrule
\textbf{Qwen3-4B}                 &       &       &        &               &               &               &         \\
Correct Rollouts Only         & 85.47 & 71.67 & 25.56  & 54.67         & 44.00         & 38.60         & 53.33   \\
Wrong Rollouts Only           & 91.47 & 80.00 & 36.67  & 53.17         & 47.50         & 44.36         & \textbf{58.86}   \\
All Rollouts            & 88.67 & 81.67 & 36.67  & 51.83         & 48.17         & 41.91         & 58.15   \\
\midrule
\textbf{Qwen3-8B}                 &       &       &        &               &               &               &         \\
Correct Rollouts Only         & 88.13 & 83.33 & 35.56  & 52.50         & 47.17         & 42.52         & 58.20   \\
Wrong Rollouts Only           & 92.07 & 82.50 & 47.78  & 56.00         & 49.17         & 44.73         & \textbf{62.04}   \\
All Rollouts             & 90.67 & 77.50 & 38.89  & 53.00         & 50.17         & 42.52         & 58.79  \\
\bottomrule
\end{tabular}
}
\caption{Ablation on symmetric versus asymmetric application of \method{} weights. Applying the saliency signal exclusively to wrong rollouts yields the highest peak accuracy across both the 4B and 8B architectures. Applying \method{} symmetrically across all rollouts provides a competitive and more stable learning signal.}
\label{tab:symmetry_ablation}
\end{table*}

\subsubsection{Symmetric vs. Asymmetric Weighting}
\label{sec:asymmetric}

To evaluate the impact of saliency weighting on different reward polarities, we ablate the application of \method{} across correct and incorrect completions. Table \ref{tab:symmetry_ablation} compares the performance of Qwen3-4B and Qwen3-8B when the saliency signal is applied symmetrically (to all rollouts) versus asymmetrically (exclusively to correct or wrong rollouts).

Applying the weights only to wrong rollouts gives the highest peak accuracy: 58.86\% on the 4B model and 62.04\% on the 8B model. In particular, this asymmetric penalty drives gains on complex tasks like AIME24 (reaching 47.78\% for Qwen3-8B). Conversely, applying the weights only to correct rollouts consistently underperforms, indicating that positive reinforcement alone is insufficient for effective credit assignment.

The symmetric version is the cleanest formulation. However, the strong result of the "wrong rollouts only" setting is informative. This empirical result suggests that \method{} is more effective at identifying logical errors than it is at finding pivotal correct steps. As corroborated by our positional analysis (Section \ref{sec:positional_analysis}), correct rollouts produce a diffuse saliency signal across intermediate steps, whereas flawed derivations in wrong rollouts generate sharp, highly concentrated saliency weights. This suggests that negative reinforcement is especially useful for correcting flawed reasoning paths.

\subsection{Positional Analysis of Saliency Weights}
\label{sec:positional_analysis}

\begin{figure}[h]
    \centering
    \fbox{ \includegraphics[width=0.7\linewidth]{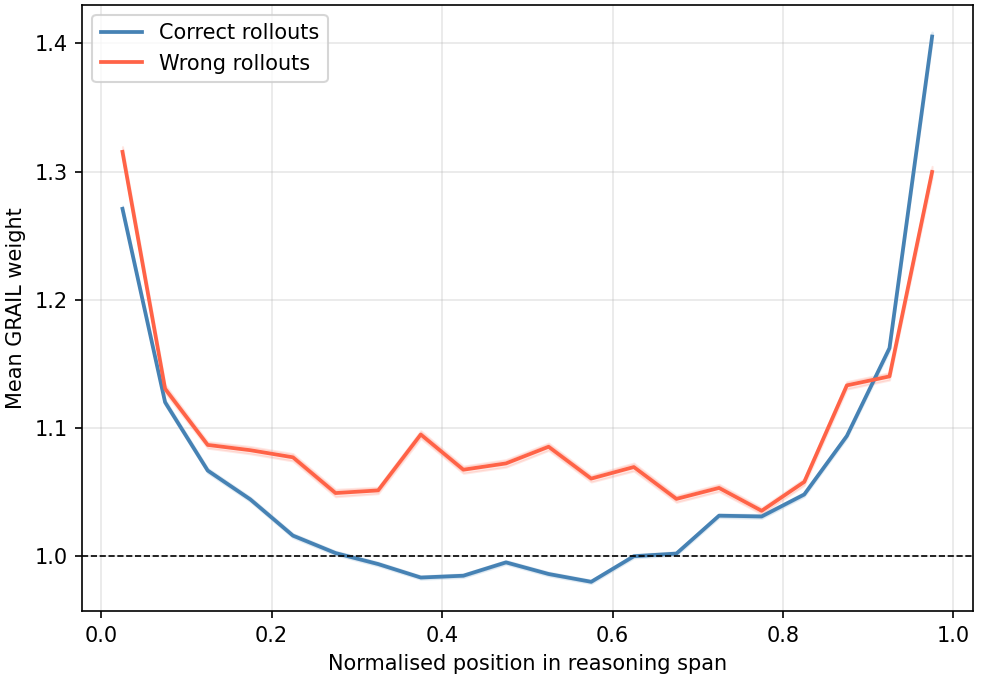}}
    \caption{Mean \method{} weights across normalised positions within the reasoning span.}
    \label{fig:positional_ablation}
\end{figure}

To understand how \method{} distributes credit across the reasoning chain, we analyse the spatial distribution of saliency weights. For each completion, we normalise the positional indices of the reasoning span to a continuous range of $[0, 1]$. We then aggregate the mean \method{} weight at each normalised position across the dataset, explicitly segregating the trajectories into correct and wrong rollouts.

The resulting distribution in Figure \ref{fig:positional_ablation} has a clear "U-shape" across completions. Saliency weights peak significantly at the beginning (positions $< 0.1$) and the end (positions $> 0.9$) of the reasoning span. This boundary concentration is consistent with the primacy-recency bias documented in transformer attention: \citet{liu2024lost} show that language models disproportionately attend to tokens at the boundaries of long contexts, with middle tokens receiving systematically weaker attention density. Under the causal attention mask, initial premise tokens are attended to by every subsequent token in the sequence, while final concluding tokens directly precede the answer markers that define $\mathcal{L}_\text{ans}$. Both positions thus accumulate high gradient signal regardless of their semantic content, producing the observed boundary peaks.

The more diagnostically significant difference lies in the intermediate reasoning span (normalised positions $0.2$ to $0.8$). For correct rollouts, the mean \method{} weight drops to, or slightly below, the neutral baseline ($w \approx 1.0$), producing a diffuse and approximately uniform advantage signal. This is consistent with the intuition that a correct derivation distributes its logical contribution broadly, as no single intermediate step is markedly more responsible for the final answer than others. In contrast, wrong rollouts consistently maintain a higher mean weight across this entire intermediate span. We hypothesise that this concentration reflects locations of logical failure: when an error is introduced mid-derivation, the gradient of $\mathcal{L}_\text{ans}$ with respect to the input embeddings becomes concentrated on the tokens surrounding that failure, since the attention mechanism routes information from that erroneous step into all subsequent hidden states and ultimately into the answer tokens. This asymmetry could explain the empirical finding from Section~\ref{sec:asymmetric}, where penalising wrong rollouts with the GRAIL signal is substantially more effective than reinforcing correct ones, since the saliency weights are most discriminative precisely where they are most needed.

\section{Conclusion}

In this work, we addressed the limitation of uniform credit assignment in Group Relative Policy Optimization (\grpo{}) for mathematical reasoning. We hypothesised that broadcasting a sequence-level advantage equally to all tokens dilutes the policy gradient, failing to distinguish between critical logical inferences and flawed intermediate derivations. To resolve this, we introduced Gradient-Reweighted Advantage (\method{}), a loss reweighting mechanism that derives token-specific importance weights. Extensive evaluations across the five models demonstrate that \method{} consistently elevates reasoning performance, yielding substantial gains on complex, long-horizon tasks. Crucially, as revealed by our positional analysis, \method{} achieves this by assigning higher weights to intermediate tokens linked to logical errors. This focuses the learning signal where the policy needs correction.

\method{} provides a self-contained way to improve fine-grained reasoning alignment in large language models. While currently evaluated on mathematical problem-solving, the nature of this intrinsic gradient-weighting mechanism carries broad implications for the verifiable reinforcement learning paradigm. Future work may explore the application of this saliency framework beyond mathematics, extending its utility into other domains that require verifiable trajectories. \method{} grounds the weighting signal in the model's own gradients, which makes it a simple path for improving verifiable RL.

\section*{Limitations}

While Gradient-Reweighted Advantage (\method{}) provides an effective intrinsic learning signal, the gradient-activation product relies on a first-order approximation of influence. This approach successfully captures the linear sensitivity of the final answer to each token's input representation, but it does not account for higher-order interactions, such as tokens that only become critical in combination with others or counterfactual contributions. Although advanced saliency methods like Integrated Gradients could provide a more theoretically rigorous estimator, implementing them during continuous reinforcement learning would require multiple forward and backward passes per rollout, making them computationally prohibitive.

\bibliographystyle{assets/plainnat}
\bibliography{custom}

\appendix
\newpage
\section{Hyperparameters}
\label{sec:hyperparam}

\subsection{Training Hyperparameters}

\begin{table}[h]
\centering
\resizebox{0.4\columnwidth}{!}{%
\begin{tabular}{lc}
\toprule
\textbf{Hyperparameter} & \textbf{Value} \\
\midrule
\multicolumn{2}{c}{\textit{Optimization \& Training}} \\
\midrule
Precision & \texttt{bfloat16} \\
Attention Implementation & Flash Attention 2 \\
Optimizer & Fused AdamW \\
Learning Rate & $1.0 \times 10^{-6}$ \\
Learning Rate Scheduler & Constant \\
Warmup Steps & 20 \\
Total Train Batch Size & 256 \\
Max Steps & 200 \\
Max Gradient Norm & 1.0 \\
Random Seed & 42 \\
\midrule
\multicolumn{2}{c}{\textit{\grpo{} \& Rollout Generation}} \\
\midrule
Loss Type & \texttt{dr\_grpo} \\
Group Size ($|G|$ / \texttt{num\_generations}) & 8 \\
GRPO Iterations (\texttt{num\_iterations}) & 4 \\
KL Penalty ($\beta$) & 0.0 \\
Clip Range ($\epsilon$) & 0.2 \\
Upper Clip Range ($\epsilon_{\text{high}}$) & 0.28 \\
Max Completion Length & 4096 \\
Sampling Temperature & 1.0 \\
Top-$p$ & 1.0 \\
Top-$k$ & 50 \\
Repetition Penalty & 1.0 \\
Mask Truncated Completions & True \\
\midrule
\multicolumn{2}{c}{\textit{Gradient-Reweighted Advantage (\method{})}} \\
\midrule
Neutral Baseline Weight ($w_{\text{mean}}$) & 1.0 \\
Standard Deviation ($\sigma_w$) & 0.5 \\
Minimum Weight Bound ($w_{\min}$) & 0.5 \\
Maximum Weight Bound ($w_{\max}$) & 5.0 \\
\bottomrule
\end{tabular}
}
\caption{Hyperparameters used for training Qwen3 and OctoThinker models via the Hugging Face \texttt{trl} library.}
\label{tab:hyperparameters}
\end{table}

All models are fine-tuned using the open-source \texttt{trl} library from Hugging Face \citep{vonwerra2020trl}. The GRPO objective is configured to utilize the Dr.\ GRPO loss formulation (\texttt{dr\_grpo}), with the KL divergence penalty ($\beta$) set to 0. A comprehensive list of the hyperparameters used across our training runs is provided in Table \ref{tab:hyperparameters}.

\subsection{Evaluation Hyperparameters}

\begin{table}[h]
\centering
\resizebox{0.3\columnwidth}{!}{%
\begin{tabular}{lc}
\toprule
\textbf{Hyperparameter} & \textbf{Value} \\
\midrule
Number of Generations & 3 \\
Sampling Temperature & 0.6 \\
Top-$p$ & 0.95 \\
Top-$k$ & 20 \\
\bottomrule
\end{tabular}
}
\caption{Sampling hyperparameters utilized during the benchmark evaluation phase.}
\label{tab:eval_hyperparameters}
\end{table}

During the evaluation phase across all six mathematical reasoning benchmarks, we maintain a consistent sampling and verification protocol to ensure robust and fair comparisons. For each problem, the model generates multiple candidate solutions. The final mathematical answer is extracted from the reasoning trace by parsing the contents of the \texttt{\textbackslash boxed\{\}}. We then utilize the \texttt{math-verify} library to evaluate the mathematical equivalence between the extracted prediction and the dataset's ground truth, which accounts for varied algebraic and symbolic representations \citep{math-verify}.

The specific sampling parameters used to generate the evaluation rollouts are detailed in Table \ref{tab:eval_hyperparameters}.

\subsection{Prompt}

To elicit structured, step-by-step reasoning and ensure the final answers are reliably formatted for automated extraction, we wrap all mathematical queries using a standardized zero-shot template. During both the training rollout generation and the final evaluation phases, the input query $q$ is formatted as follows:

\begin{quote}
\texttt{\{Question\}. Please reason step by step, and put your final answer within \textbackslash boxed\{\}.}
\end{quote}

This explicit structural constraint guarantees that the generated reasoning traces are compatible with the \texttt{math-verify} extraction pipeline, while also providing a consistent delimiter for the GRAIL mechanism to anchor its positional weight corrections.

\section{Computational Efficiency and Training Time}
\label{sec:efficiency}

Calculating the intrinsic token-level weights in \method{}  introduces a computational overhead. Specifically, deriving the saliency scores requires an additional backward pass through the attention layers to the input embeddings prior to the policy update step.

To quantify this overhead, we report the total wall-clock training time for both the standard \grpo{} baseline and \method{} across our evaluated models. All training runs were conducted on a single compute node equipped with 4 $\times$ NVIDIA H200 GPUs.  As detailed in Table \ref{tab:training_time}, \method{} introduces an average computational overhead of approximately 50\% to 60\% over \grpo{}.

\begin{table}[h]
\centering
\resizebox{0.5\columnwidth}{!}{%
\begin{tabular}{lccc}
\toprule
\textbf{Model} & \textbf{GRPO} & \textbf{GRAIL} & \textbf{Overhead} \\
\midrule
OctoThinker-3B & 1h 05m & 1h 40m & +54\% \\
Qwen3-4B       & 1h 41m & 2h 30m & +49\% \\
OctoThinker-8B & 2h 05m & 3h 20m & +60\% \\
Qwen3-8B       & 2h 30m & 4h 00m & +60\% \\
\bottomrule
\end{tabular}
}
\caption{Comparison of total wall-clock training time between standard GRPO and \method{}. All models were trained for 200 steps on 4 $\times$ NVIDIA H200 GPUs.}
\label{tab:training_time}
\end{table}

\end{document}